\relax
\documentclass{article}
\usepackage{ijcai19}

\usepackage{times} 
\usepackage{helvet} 
\usepackage{courier} 
\usepackage{url} 
\usepackage{graphicx} 
\usepackage[utf8]{inputenc}

\usepackage{footnote}
\usepackage{tablefootnote}

\usepackage[para,online,flushleft]{threeparttable}

\usepackage{amsfonts,bm,color}
\usepackage{booktabs}    
\usepackage{amsopn}
\usepackage{amsmath}
\usepackage{amssymb}
\usepackage{empheq}
\usepackage{epstopdf}
\usepackage{float}
\usepackage{framed}
\usepackage{multirow}

\usepackage{adjustbox}
\usepackage{algorithm,algorithmicx}
\usepackage{algpseudocode}
\usepackage[small]{caption}

\usepackage{subcaption}

\DeclareMathAlphabet\mathbfcal{OMS}{cmsy}{b}{n}

\definecolor{Sijia_color}{rgb}{0., 0., 0.}
\definecolor{Horst_color}{rgb}{0., 0., 0.}
\definecolor{PRC}{rgb}{0., 0., 0.}

\definecolor{aquamarine}{rgb}{0.5, 1.0, 0.83}
\definecolor{orange-red}{rgb}{1.0, 0.27, 0.0}
\definecolor{gray}{rgb}{0.5, 0.5, 0.5}
\definecolor{blue}{rgb}{0.0, 0.0, 1.0}
\definecolor{fuchsia}{rgb}{1.0, 0.0, 1.0}
\definecolor{black}{rgb}{0.0, 0.0, 0.0}

\begin{document}
%
\title{How can AI Automate End-to-End Data Science?}

\author{Charu Aggarwal\and Djallel Bouneffouf\and
 Horst Samulowitz\and  Beat Buesser\and Thanh Hoang\and \\ Udayan Khurana\and
 Sijia Liu\and 
 Tejaswini Pedapati\and
 Parikshit Ram\and
 Ambrish Rawat\and \\
 Martin Wistuba \And Alexander Gray
 \\
 \affiliations
 IBM Research AI
 \emails
 \{First Name . Last Name\}@us.IBM.com
}

\maketitle

\begin{abstract}
Data science is labor-intensive and human experts are scarce but heavily involved in every aspect of it. This makes data science time consuming and restricted to experts with the resulting quality heavily dependent on their experience and skills. To make data science more accessible and scalable, we need its democratization. Automated Data Science (AutoDS) is aimed towards that goal and is emerging as an important research and business topic. We introduce and define the AutoDS challenge, followed by a proposal of a general AutoDS framework that covers existing approaches but also provides guidance for the development of new methods. We categorize and review the existing literature from multiple aspects of the problem setup and employed techniques. Then we provide several views on how AI could succeed in automating end-to-end AutoDS. We hope this survey can serve as insightful guideline for the AutoDS field and provide inspiration for future research. 
\end{abstract}

\section{Introduction and Motivation} \label{sec:intro}

\begin{figure}[tb]
    \centering
    \includegraphics[width=0.3\textwidth]{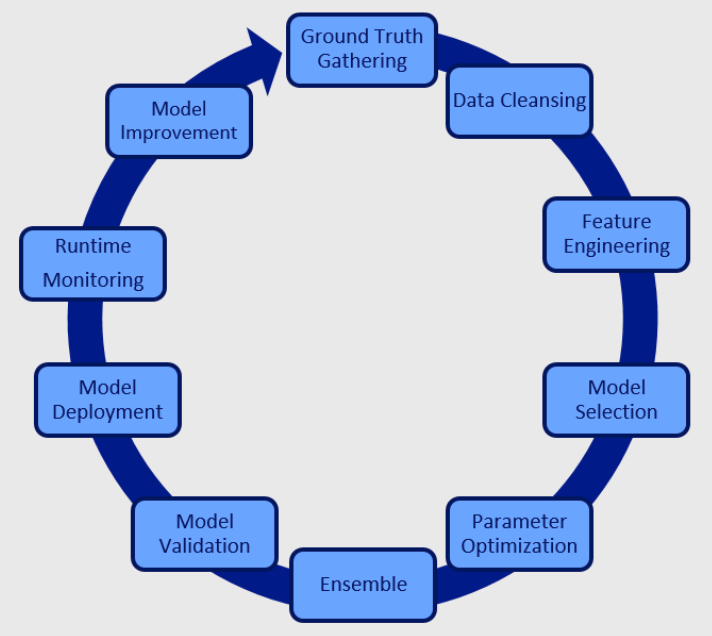}
    \caption{Steps involved in a typical data science workflow.}
    \label{fig:DS_steps}
\end{figure}

Data science covers the whole spectrum of data processing, beginning from data integration, distributed architecture, automating machine learning, data visualization, dashboards and BI, data engineering, deployment in production mode, and automated and data-driven decisions (Figure \ref{fig:DS_steps}). A key part of data science is machine learning in which the system learns from data-driven examples in order to make predictions about examples in which some of the attributes are missing. In supervised learning, this process of modeling uses fully specified examples, referred to as training data, whereas in unsupervised learning, this process is done with incompletely specified examples. Beyond these core areas, many parts of data science such as data acquisition, data integration, and data visualization (which are traditionally not considered machine learning) are essential to the effective deployment of data science and machine learning solutions. 

Much of data science requires manual intervention in the form of the choice of software (and its set up) and the overall learning pipeline; this makes the use of available resources over diverse settings very challenging. For example, machine learning often requires significant manual exploration of the input data before the data science pipelines are experimented for various tasks. This is a time consuming task, and is often the primary bottleneck in the ad~hoc effort required from the analyst. 
Given a set of machine learning tools and a set of problems, how does one choose what type of resource to deploy in a particular setting? How does one decide how to set its parameters? When training has been performed with a particular data set, how does one decide that the constructed model has now become stale? All these questions often require manual intervention from a user who must experiment with various settings in trial-and-error mode, until decisions on deployment have been made. This type of situation militates against the successful use of wide resources over many settings. A fundamental question in automated data science is whether one can automate large parts of the learning process so that key {\em meta-decisions} on model choice, (hyper)parameter tuning, and model freshness can be made with minimal user intervention. Indeed, the holy grail of being able to make machines truly intelligent is to automate large portions of machine learning that remain hand crafted to a large extent even today. Even though deep learning has already helped in automating many tasks such as feature engineering (that were earlier hand crafted in traditional machine learning), there remain significant lacunae in what is truly possible in terms of reducing the need for human intervention. 

The challenges in automating data science may arise at individual stages of the pipeline, or in the construction of the entire pipeline itself. Furthermore, each stage in the pipeline poses its own set of unique challenges -- for example, challenges in the data integration stage are quite different from those in the machine learning stage. We will discuss these challenges in the context of automated data science solutions. 
Rather than an exhaustive survey, our purpose in this paper is mainly to provide a qualitative overview and editorial outlook for the topic of end-to-end automation of data science. We begin by describing what artificial intelligence (AI) currently automates in the parts of the data science pipeline, then highlight some general approaches that could address the entire process more holistically.


\section{What Does AI Automate Today?}
Although the goal of end-to-end automated data science is still quite far, many parts of the data science pipeline have been automated with significant success. Here we provide a broad overview of many such cases.
\subsection{Automated Data Collection}
Data engineering is the aspect of data science that focuses on the {\em ground truth gathering} step with the focus on data collection and analysis. For all the questions that data scientists answer using large data sets, there have to be mechanisms for collecting and validating those relevant information. In this context, much of modern hardware technology has already automated data collection; simple transactions today such as phone usage or credit card swipes lead to data collection behind the scenes. The real issue is that there is a {\em cost} in even using the data collected unless there are ways of parsing out parts of the data relevant and useful for the application at hand. The field of {\em active learning}~\cite{activesurvey} provides methods capable of selectively collecting appropriate and relevant data for a particular application. This process integrates data collection and knowledge discovery, and automates large parts of the process to minimize manual effort.
\subsection{Automated Data Integration}
The process of integrating data from heterogeneous sources into a single, unified view
has been a significant barrier to many data science tasks --
a variety of infrastructures generate different data formats with varying levels of contamination in the raw data.
This necessitates custom ETL (Extract, Transform, Load) code for {\em data cleaning and integration}..
This process has been automated with a declarative interface \cite{kougka2018many} allowing for extendable domain-specific data models; a AI planner performs the integration, optimizing for the plan completion time.
%
This tool is designed for schema-less data querying, code reuse within specific domains, and is robust to messy unstructured data, 
demonstrated by its capability of integrating data from diverse sources such as web click-stream logs and the census.
%
However, the application of such data cleaning and integration tools have been relatively limited, and there is various open research challenges to wider automation. For example, for any given set of data sources, a truly autonomous system should be able to automatically detect the nature of the data and the ETL steps required for the application at hand.
%
%
\subsection{Automated Feature Engineering }
The use and effectiveness of different approaches to {\em feature engineering} is heavily dependent on the nature of the data and the learning task. Problems involving images, videos and texts have seen significant success with deep neural networks where the feature engineering is an integral part of the modeling step, and explicit separate feature engineering has received less attention. In contrast, tabular and relational data rely heavily on explicit feature engineering (automated or otherwise) -- it is often the most crucial step in terms of the downstream predictive performance, and consequently being the most time consuming, making it a prime candidate for automation. We categorize the automation schemes based on the type of data they are applied to:
%

\noindent
\textbf{Tabular data:} For tabular data, the basic set of features can be extended to transform the data into a feature space that benefits specific models. 
The set of transformations is usually very large and efficient automation cannot be achieved by exhaustively trying all possible transformations. One way of reducing the size of the set is by learning the effectiveness of different transformations for a given ``type'' of prediction problem on data seen in the past by the system \cite{lfe}.
%
This relies on characterizing the problem (data features, targets, objectives) and effectiveness of each transformation. For any new data, the system suggests transformations based on its learned experience. It does so by learning meta-predictors (which themselves are classification or regression models) on the historical data and statistics. The meta-learning step requires essential preprocessing of the data through quantile sketching which converts data of different shapes and sizes to a canonical format. The prominent advantage of this approach is the low runtime cost, which is due to a handful of inference steps. 

\noindent
\textbf{Relational data:} According to a recent survey among more than 8000 data scientists \cite{Kaggle2017}, 65\% are frequently working with relational data, making it the most popular data type. However, manual feature engineering with relational data is a tedious task requiring 
multiple SQL queries in an error-prone trial-and-error fashion until desired results are obtained.
Therefore, automated feature engineering for multi-relational data has recently received increasing attention. There are two basic problems: the first involves the choice of appropriate joining paths in entity relation graphs to collect relevant data for a given prediction target, and the second requires choice of the right transformations/aggregations to turn the joined tables into useful features. 
An early method learned inductive decision trees by propositionalizing relational data (transforming relational data with multiple tables into a single table with features) \cite{DBLP:conf/pkdd/KnobbeSW99}.
%
The problem has been recently revisited with Deep Feature Synthesis (DFS) \cite{DBLP:conf/dsaa/KanterV15}, but limited to numerical data. The One Button Machine (OneBM) extends support for complex transformations on non-numerical data ~\cite{DBLP:journals/corr/LamTSCMA17}. Both automated schemes are rule-based where the transformations and joining path choices are predefined based on heuristics. While choosing the right join is a computationally intractable problem, transformations can be learned from relational data using deep neural network \cite{r2n}. 
\subsection{Automated Machine Learning}
Automated Machine learning (AutoML) has received 
increasing attention, starting with hyper-parameter optimization (HPO) 
to determine the most appropriate parameters for a ML model (for example, the number of trees in a random forest), 
to automating selection of a ML pipeline (such as feature transformation \&  selection combined with predictive modeling). 
AutoML automates the {\em model selection}, {\em (hyper)parameter optimization}, and even {\em ensembling} steps of the data science workflow, addressing a wide range of difficult technical challenges ranging from HPO, automated feature engineering, to neural network design. The neural network paradigm has itself been a fillip to the development of automated data science by enabling automated feature engineering to a large extent, and has provided successful end-to-end solutions in specific domains like machine translation~\cite{wu2016google} which traditionally relied on large amounts of hand-crafted features. 
Nevertheless, the move to fully automated systems has been incomplete because they continue to require human-centric tuning of large numbers of parameters or design choices. 
AutoML bridges this additional gap to a large extent; in addition, newer technologies such as reinforcement learning can play a significant role. 
Auto-WEKA~\cite{autoweka1} and Auto-sklearn \cite{feurer2015efficient} are the main representatives for solving AutoML by so-called sequential parameter optimization. Both apply Bayesian Optimization to find useful ML pipelines.
Auto-sklearn improves upon Auto-WEKA by utilizing {\em meta-learning} \cite{vanschoren2018meta} and {\em ensembling}.
\subsection{Visualizations and Decision Making}
At the end of the day, the results of machine learning are often fed into visualization and/or decision making tools. 
This is a particularly tricky part of the process, since the nature of the visualizations and decisions are highly application-specific. This makes automation less feasible. Nevertheless, progress has been made in some partial respects. 

The ability to create good visualizations became a must-have skill for all data analysts. In current data visualization tools, users need to know their data well in order to create good visualizations. However, the users need tools to automatically recommend visualizations rather than hand-craft highly customized tools. 
The authors in \cite{luo2018deepeye} propose a system for automatic data visualization that tackles the problem of Visualization recognition where given a visualization, it provides a prediction of whether it is “good” or “bad”. This is achieved by training a binary classifier to model the quality of visualization. They also study the Visualization ranking problem, which provides a relative ordering of two given visualizations. This is achieved with a supervised learning-to-rank model, although expert knowledge is also considered with the use of expert rules. Therefore, the approach is not a fully automated system yet, and it uses human criteria in the form of expert rules. Nevertheless, this still provides a modicum of automation in the drive towards automated visual systems. 

Beyond visualization, the final goal of data science is to support decision making. 
Automated Business Intelligence systems are software applications that utilize automated processes in order to extract actionable organizational knowledge. Authors in \cite{Soper2005} proposes an architecture to guide the development of such systems and in so doing outlines a feasible approach by which organizations can adopt them in support of their strategic decision making processes. The goal is to gain a competitive advantage by utilizing information garnered from web sources to inform corporate decision making. Similar to \cite{Soper2005}, the authors in \cite{nagy2009informatics} proposed to automate the extraction, processing, and display of indicators provide useful and current data for operational meetings. The feasibility of extracting specific metrics from information systems was evaluated as part of a longer-term effort to build a business intelligence architecture. Analytics were performed on the data, a process that generated indicators in a dynamic Web-based graphical environment that proved valuable in discussion and root cause analysis. However, this type of decision making is still not fully automated. A key aspect that distinguishes it from truly intelligent (human-like) systems is the trial-and-error process that is endemic to all forms of intelligent decision making. This will be the topic of discussion in the next section. 

\section{How can AI Automate End-to-End Data Science?}
Much of what AI automates today remain as parts of the data science pipeline. However, to be able to work from raw sensory inputs to final decisions is a key challenge that is required for a high level of automation. This forms the basic goal of {\em end-to-end} data science, as opposed to the automation of individual parts of the data science process, discussed in the previous sections.  Here we highlight general frameworks and approaches which offer possible avenues for considering more holistic automation.


\subsection{Reinforcement Learning} \label{sec:DRL}
The main challenge that arises in building fully automated systems at the level of humans is the fact that the construction of a machine learning system requires a large number of design choices, and the specific {\em combination} of these design choices can regulate the effectiveness of the system at hand. In many cases, these decisions need to be made {\em sequentially}. For example, if one is to construct a neural network for a particular task, then the choice of the number of layers naturally precedes the choice of the number of units in each layer.
Humans are naturally prone to experimenting with these large numbers of decision choices, and are often able to construct a reasonably accurate system with a relatively modest number of iterations of trial and error. Therefore, creating systems that {\em automatically perform trial and error, and learn from successes and failures is the key to creating a truly intelligent automated system.} This type of setting is naturally the domain of {\em reinforcement learning} in which a system can learn from the success and failure of trials~\cite{sutton98}. Reinforcement learning is used extensively for video games~\cite{mnih2013,NoothigattuBMCM19}, in which one makes a set of sequential decisions in order to win virtual rewards in the form of game points or victories, and the success of a particular set of decisions can be easily judged. It is noteworthy that many of these solutions work with raw sensory inputs (e.g., pixels) and provide the final decisions as the result, which is a high level of end-to-end automation. This situation applies perfectly to the automated machine learning paradigm where the success of a particular design choice can be easily evaluated. 

Reinforcement learning is also used through the bandit framework \cite{AllesiardoFB14,LinBCR18,bouneffouf2016multi,BalakrishnanBMR19,Bouneffouf0SW19}. 

For instance, the authors in \cite{bouneffouf2012contextual,bouneffouf2012exploration,bouneffouf2013contextual,bouneffouf2012hybrid,bouneffouf2014contextual} introduce an algorithms that tackles this dilemma in Context-Based Information Retrieval, Context-Based recommender system and in active learning. It is based on dynamic exploration/exploitation and it can adaptively balance the two aspects by deciding which situation is most relevant for exploration or exploitation. Using combinatorial bandit frameworks, authors of \cite{BouneffoufRCF17} tackle the online feature selection problem by addressing the combinatorial optimization problem in the stochastic bandit setting with bandit feedback, utilizing the Thompson Sampling algorithm.

\subsection{Deep Learning} \label{sec:ADL}
The problem of Automated Deep Learning (ADL) boils down to designing a strategy that given a dataset and a task along with some constraints, yields a well-trained deep learning model that can be used for solving the task.
Designing an optimizer that given a search space, looks for an optimal architecture, forms the key component of ADL methods.
The current trends in ADL methods explore a variety of approaches to solve this optimization problem, most of which build upon the well established theories in reinforcement learning (RL) and evolutionary algorithms (EA). It should come as no surprise that both of these ideas borrow heavily from the success of the  biological paradigm.

As a specific example, consider the case where one wants to learn the  best neural architecture for a particular data domain in a problem-specific way. The RL-based optimizers involve learning a controller which is trained to output actions that lead to architecture encodings which correspond to high-performance deep learning models.
\cite{baker2016designing} and \cite{zoph2016neural} were one of the first works to explore RL-based approaches for architecture search.
Both these works build networks by sequentially choosing its layers and its set of operations.
The validation accuracy of the final trained network is used as a reward to update the controller.
\cite{Cai2018_Efficient} take a different approach.
Starting from a given architecture, they learn how to modify it in order to improve its performance.
In this context, an interesting reinforcement learning system is proposed in~\cite{zoph2016neural}, which creates an optimal convolutional neural network architecture for image classification. The convolutional neural network whose architecture is to be constructed is considered the {\em child network}, whereas the neural network that creates the architecture is referred to as the {\em controller network}. The controller network is a recurrent neural network. The controller network is used to make decisions about the parameters of the child network, and correct decisions are rewarded by higher accuracy of prediction. These types of controller-child combinations are coupled with a boiler-plate reinforcement learning algorithm such as {\em policy gradients}~\cite{williams1992simple}. The basic idea is that the controller network (together with the reinforcement learning algorithm) experiments with the parameters of the child network and encodes correct decisions within the parameters of the controller network. The controller network is essentially a {\em policy network} that outputs probabilities of specific decisions about the design of the neural architecture. This broad approach is very similar to how a human experiments with the parameters of a neural network and learns from the experiences obtained with the resulting performance. Other variants of techniques that use reinforcement learning to design neural architectures are discussed in~\cite{baker2016designing}. The main drawback of the reinforcement learning paradigm is that it is computationally expensive, and it can sometimes be difficult to repeatedly train systems with different sets of parameters in order to judge the success of a particular setting.


A key bottleneck in these methods that the evaluation of an architecture requires expensive and repeated training.
This problem can be alleviated by reusing parameters of smaller or shallower networks with function-preserving operations~\cite{Chen2015_Net2Net}.
The resulting ``warm-start'' model  requires fewer epochs to train a new network~\cite{Cai2018_Efficient,Wistuba2018_Deep}.
In another line of work, a large common network is trained, and it is hypothesized that sampling architectures from this large network (either uniformly~\cite{Bender2018_Understanding} or according to an optimizer~\cite{Pham2018_ENAS})  will yield a useful ranking of the sampled architectures.
Some works relax the assumption that the choice of operations has to be discrete and learn a parameterized architecture choice jointly with model parameters~\cite{Liu2018_DARTS,Xie2019_SNAS}.  Other works extrapolate the learning curve of a network~\cite{Baker2017_Accelerating} by predicting whether a training run will improve the current best solution (and terminating the run early).

It is worth noting that high accuracy is often not the only objective of a deep learning model.
Often additional practical constraints like number of parameters or inference time need to be incorporated.
This task is either solved by aggregating all objective functions to a single one and solving it with standard ADL optimizers~\cite{Hsu2018_MONAS} or approaches which search for pareto-optimal solutions~\cite{Elsken2019_Efficient}.

Search spaces are a useful component in defining the ```scope'' within which an optimizer will look for an optimal architecture. For example, in the convolutional neural network setting, does one use a conventional network like VGG as the ``base'' model, or does one use a modern skip-connection-based {\em ResNet} as the base model?  The nature of the search space in the latter is more complex. 
The search spaces have progressively become complex in their definitions over the course of developments in automated deep learning approaches.
The cell-based search space, initially proposed by~\cite{Zoph2018_Learning}, enables an easy transfer of deep learning models across datasets and tasks.
The task for the ADL optimizer reduces to select cells which are stacked to derive the final architecture.
A cell is built as a combination of blocks, each block has two inputs and two corresponding operations and a combination operator to output a new state.
The flexibility of picking input states allows for learning artifacts like parallel convolutions, branches, and skip-connections.

Architecture search is not the only component of deep learning modeling pipeline that has benefited from automation.
Additional components like search for heuristics of optimization methods~\cite{Bello2017_Neural}, a search for suitable activation functions~\cite{Ramachandran2018_Searching}, storing few experiences \cite{RiemerKBF19}, and automated data augmentation~\cite{Cubuk2018_AutoAugment} has also been investigated for automation in some of more recent works.

\subsection{Black-Box Optimization}\label{sec:bbopt}
One of the major challenges in designing automated learning system is its \textit{black-box} optimization nature when explicit expressions of the gradients are difficult to obtain. For example the problems of architecture search and feature engineering do not offer a continuous loss function that can be optimized with gradient descent. The only mode of interaction with the learning system is by submitting inputs and receiving  feedback. In AutoDS, the commonly-used black-box optimization methods include Bayesian optimization (BO) \cite{shahriari2016taking}, zeroth-order (ZO) optimization \cite{nesterov2015random}, derivative-free trust region method (DF-TRM) \cite{conn2009global}. 
BO has been largely used in AutoDS for 
 the model selection and hyper-parameter
tuning tasks \cite{autoweka1}. However, it suffers from poor scalability with increasing dimensionality. 
ZO optimization mimics first-order descent-type methods 
 and have been applied in AutoDS for solving multi-armed bandit problems~\cite{agarwal2010optimal}. In contrast to BO, ZO optimization has provably favorable convergence. However, it suffers from high query complexity and the smoothness requirement in the objective function. 
DF-TRM approximates the black-box function with a parametric surrogate function (such as linear or quadratic functions) fit on the available function evaluations, which allows for efficient constrained optimization on the surrogate
\cite{conn2009global,conn2009introduction,ChoromanskaCKLR19,SijiaADMM2019}. For the continuous variables they are restricted to a neighborhood of the current point (the trust region) and for the discrete variables the number of changes is restricted. The size of the trust region corrects for the discrepancy between the parametric surrogate and the black-box.
%
DF-TRM is computationally intensive, impeding its widespread use compared to BO and ZO optimization.
At this point its also worthwhile mentioning that one can combine multiple AI technologies such as leveraging AI planning in this context~\cite{Biem2015TowardsCA} to tackle the various challenges.

\subsection{Evolutionary Algorithms} \label{sec:ea}
Reinforcement learning methods follow the ubiquitous biological paradigm of reward-based trial and error in order to create automated systems for AI. Another natural paradigm that copies  biological principles is that of evolutionary algorithms. These methods encode solutions to problems (e.g., neural architectures) as codes that are akin to chromosomes in the biological DNA. Just as biological organisms evolve and become intelligent through (Darwinian) reward-driven trial-and-error, evolutionary algorithms (EA) use a Darwinian approach of selecting highly-performing architectures and recombining their best characteristics to explore the search space. It is not difficult to see that this is a rough simulation of how the biological brain has evolved over the millenia. 
EA-based optimizers follow a standard optimization strategy wherein a population of networks is maintained and a set of evolution steps are carried out until termination.
This traditionally involves 1) selecting ``parents'' (i.e., highly performing architectures) in a ``Darwinian'' way 2) applying mutation and recombination operations to create new ``individuals'' 3) evaluating the ``fitness'' of the recombined individuals. This is followed by repeating the Darwinian process of selecting the most fit survivors of the population and recombining them again.
Evolutionary approaches span a highly diverse set of ADL methods each of which vary in their definitions of encodings, set of mutations and selection strategies, notable ones being~\cite{Real2017_Large,Real2019_Aging}.
The main challenge arises from the large search space and over which one must optimize the population. Indeed, the success of biological evolution is owed to the fact that the evolution of the ecosystem can be viewed as a massively (but loosely) parallel process that has occurred over hundreds of millions of years over millions of species. We have nothing approaching  that kind of computational power today. 
\subsection{Meta-Learning} \label{sec:ul}
 A truly automated system can make reward-driven decisions, just like any intelligent organism. Although this goal is achieved to some extent by reinforcement learning, the main problem with reinforcement learning solutions is that they require massive amounts of data, and are therefore mostly good for learning in artificial settings like board/video games, where one can generate unlimited amounts of data with self-play. Similarly, although reinforcement learning robots can learn to work on their own in virtual simulators relatively easily, this is much harder to generalize to physical robots, where the speed of data collection is constrained by the limitations of the number of tasks a physical   robot can perform in a specific period of time, and also by the fragility of physical robots to the consequences of ``bad'' trials. On the other hand, most humans can learn from relatively limited amounts of data. For example, a child does not need too many examples of a toy truck to learn what it is, and can easily generalize the idea to real-world trucks of completely different shape, size, and color.
 This difference between humans and automated learning systems can be partially explained by the fact that the former is far more powerful in terms of its ability to perform unsupervised learning. Humans take in massive amounts of sensory input, which is processed in an unsupervised way in order to perform continuous learning over time. This unsupervised learning can be viewed as a continuous process of massive ``pre-training'' that makes it easier to learn specific tasks with a smaller amount of data. Furthermore, much of the unsupervised {\em and} supervised learning is encoded in the highly ``regularized'' neural structure of the brain, which has evolved over millions of years and is inherited from one generation to the next. This inheritance from one generation to the next is a form of {\em transfer learning}~\cite{qiangtransfer}, which is required to be incorporated more seamlessly in automated systems than is possible with the highly customized systems available today.   Ideas surrounding learning over many past problems in order to learn more efficiently for the current problem is often called {\em meta-learning} or {\em learning to learn}~\cite{thrun}.

\section{Outlook and Perspective}
Whatever AI approach will succeed in automating end-to-end data science it will need to overcome a number of challenges:
\noindent
\textbf{Generalization.} The first is the issue of overfitting that is further amplified by automation since a machine can essentially fine-tune the flow until generalization is non-existent. In some cases,  the original goals of the application are compromised due to an automated system, such as an RL system, overfitting to difficult-to-specify rewards. While generalization has been widely studied, the implications of fully automated systems in this respect are not quite as well studied~\cite{dwork2015}.

\noindent
\textbf{Safety.}  An alarming observation in this respect correspond to safety issues-- RL systems regularly learn cheats and hacks in video games that were not originally intended. It is not useful to have  a robot that makes messes and then cleans them up-- or, more darkly, a robot nurse with similar behavior. 

\noindent
\textbf{Deep vs Non-Deep Learning.}  Many of the ideas developed solely in the context of deep learning can be applied more generally to data science with any ML method, and vice versa.  We expect and hope to see more transfer between these bodies of work.
In some data modalities, such as images, what might otherwise be considered data preparation transformations (at least certain ones) are handled by layers in the network itself, blending data preparation and model into one {\it end-to-end} training process.  This general idea can be extended beyond deep learning.  Likewise, in practice, not all of the data preparation needed for practical problems is always effectively captured in network layers, and thus the kinds of treatments of the data preparation transformations outside of deep learning can be applied to systems that employ deep learning as well.

\noindent
\textbf{Domain Knowledge.}
It is notable that this survey has given little discussion to the notion of domain knowledge, which is so important to most data scientists.  Domain knowledge can range from simple derived features such as the body mass  index (BMI) of a person, to encoding the syntax of a language for an NLP application.
While it is an article of faith among data scientists that   data science solutions  benefit from domain knowledge, its role in fully automated data science remains an open question. A key observation is that  its track record  has been mixed in more advanced forms of AI.  With limited data, domain knowledge has indeed been extremely useful as a natural regularizer. However, with increased  data, a surprising observation has been that completely data-driven learners (with zero domain knowledge) have consistently outperformed systems with encoded domain knowledge. For example, machine translators with zero knowledge of syntax now routinely outperform domain rule-based machine translators, and chess learners with zero knowledge (e.g., {\em AlphaZero}) now routinely outperform chess programs like {\em Stockfish} in which chess grandmasters have spent years in fine tuning the evaluation function.  This is simply due to the fact that domain knowledge can be a ``biased'' upper-bound to the intelligence we expect the system to eventually have. To give a biological analogy, a child can learn a lot from the domain knowledge given by parents, but it is a poor substitute to what the child can learn through their own experiences.  There is also no common-sense reasoning in a machine setting --- let alone reasoning at an domain expert level. However, there will always be some situations in which domain knowledge is useful.  Early work already exists~\cite{ruslan18}, but it still has a long way to go. 

\noindent
\textbf{Lifelong learning and unsupervised learning.}  Another is that of life-long learning - while various works exists on this topic already it is incredibly complex to automatically facilitate life-long learning. One reason is the missing taxonomy of data. For instance, when the predictions of the model influence the newly incoming data it is often not valid to use those data points for retraining. 
Unsupervised learning remains a significant challenge, because it is hard for systems to know which parts of the massive amounts of unsupervised data will be useful in future applications. 
To a large extent, it is expected that truly automated and intelligent systems will be obtained by combining reinforcement learning, unsupervised learning, and transfer learning in a way that is not fully understood today. While reinforcement learning systems require massive amounts of data, this problem will be alleviated by unsupervised learning over large periods of time that can take in massive amounts of data, and learn the portions that can obtain reinforcement rewards. At the same time, transfer learning will be required in order to inherit the generalizable knowledge over different situations in much the same way as humans inherit the highly regularized structure of their neurons across generations. In this sense, transfer learning can be viewed as a simpler and faster alternative to what one tries to achieve with the use of evolutionary algorithms.  This ability to combine unsupervised learning, supervised learning, and reinforcement learning in a reward-driven context has remained the most important problem in artificial intelligence in recent years. Recent successes in unsupervised learning, such as the ability to realistically replicate intricate data objects like images with the use of {\em generative adversarial networks} has brought significant advancements to unsupervised learning --- however, a complete and seamless integration of supervised, unsupervised, and reinforcement learning remains elusive.

\noindent
\textbf{Computational cost.}  The single largest impediment to automated data science is the computational aspect of it. One must view the limited success of automated data science by comparing it to what biological organisms have benefited from-- we have nothing close to the computational power that simulates the massively parallel ``computations'' that have occurred in the (implicit) ``computational'' process of biological evolution over  billions of organisms. One should view our relative successes and failures to biological learning and automation  from the perspective of the humbling limitations we work with today.   However, with new hardware paradigms on the horizon, such as quantum computing,  it is difficult to know how much progress we will make along these lines--- the only certainty is that one should expect the unexpected. 


 \begin{footnotesize}
\bibliographystyle{ijcai19}
\bibliography{condensed_refs} 
 \end{footnotesize}

\end{document}